\documentclass[5p,times,twocolumn]{elsarticle}
\usepackage{amssymb}
\journal{Computer Methods and Programs in Biomedicine, Volume 214, February 2022, 106584}

\begin{document}

\begin{frontmatter}
\title{Explanation of Machine Learning Models Using Shapley Additive Explanation and Application for Real Data in Hospital}

\author[1]{Yasunobu Nohara}
\author[2]{Koutarou Matsumoto}
\author[3]{Hidehisa Soejima}
\author[4]{Naoki Nakashima}

\affiliation[1]{organization={Kumamoto University}, city={Kumamoto}, country={JAPAN}}
\affiliation[2]{organization={Kurume University}, city={Fukuoka}, country={JAPAN}}
\affiliation[3]{organization={Saiseikai Kumamoto Hospital}, city={Kumamoto}, country={JAPAN}}
\affiliation[4]{organization={Kyushu University Hospital}, city={Fukuoka}, country={JAPAN}}

\begin{abstract}
When using machine learning techniques in decision-making processes, the interpretability of the models is important. In the present paper, we adopted the Shapley additive explanation (SHAP), which is based on fair profit allocation among many stakeholders depending on their contribution, for interpreting a gradient-boosting decision tree model using hospital data.
For better interpretability, we propose two novel techniques as follows: (1) a new metric of feature importance using SHAP and (2) a technique termed feature packing, which packs multiple similar features into one grouped feature to allow an easier understanding of the model without reconstruction of the model.
We then compared the explanation results between the SHAP framework and existing methods. In addition, we showed how the A/G ratio works as an important prognostic factor for cerebral infarction using our hospital data and proposed techniques.
\end{abstract}

\end{frontmatter}

\section{Introduction}
In recent years, remarkable breakthroughs have been achieved in machine learning technology, as typified by deep neural networks. Such technologies are expected to be used for decision-making in medical fields. In decision-making, it is important to recognize why decisions are made. Although complex machine learning models such as deep learning and ensemble models can achieve high accuracy, they are more difficult to interpret than simple models such as a linear model.
To provide interpretability, tree ensemble machine learning algorithm such as the Random Forest and Gradient Boosting Decision Tree (GBDT)~\cite{friedman2001greedy} can provide feature importance, which is each feature's contribution to the outcome. However, Lundberg et al. pointed out that popular feature importance values such as gain are inconsistent~\cite{lundberg2018consistent}. For addressing these problems, Lundberg et al. proposed using the SHapley Additive exPlanation (SHAP). Using the idea of the Shapley value, which is a fair profit allocation among many stakeholders depending on their contribution, SHAP represents the outcome as the sum of each feature contribution calculated as the Shapley value.
SHAP values have proved to be consistent and SHAP summary plots provide a useful overview of the model.

In this paper, we adopted the SHAP method for interpreting GBDT model constructed using real data from a hospital.
For better interpretability, we propose two novel techniques.
One is a new metric of feature importance using SHAP. Interpretation using our metric is exactly the same as that of the generalized linear model.
The other technique is feature packing, which packs multiple similar features into one grouped feature.
Our new metric of feature importance is useful for selecting similar features, whereas feature packing allows easier understanding of a model without reconstruction of the model while maintaining accuracy.
Here, we constructed a prognosis predictor using cerebral infarction data from our hospital. We then compared the explanation results between SHAP and the existing gain method.
We uncovered the underlying relationships between features and outcome using our techniques and found unknown prognosis factors.

\section{Background}
\subsection{Generalized Linear Model}
The generalized linear regression model (GLM), such as a logistic regression model, is widely used in medical studies
because it allows interpretation of the effect of explanatory variable $x_i$ for outcome $y$ using a coefficient.
In the logistic regression model, the log-odds of the probability $p$ is a linear combination of the explanatory variables, also called features $x_i \in X_i$ and coefficient $a_i$, and given as follows:
\begin{equation}
\log \frac{p}{1-p}=b+\sum_{i=1}^{K} a_i x_i
\label{glm}
\end{equation}
, where $b$ is a constant term and $K$ is the number of features.
If we increase $x_i$ by 1, the log-odds increases by $a_i$. If $a_i>0$, a larger $x_i$ has a positive impact on the outcome, while a negative $a_i$ means a negative impact of a larger $x_i$.

If all features are standardized, i.e., $\forall i, E(X_i)=0$ and $Var(X_i)=1$, the estimated coefficients are called standardized coefficients or beta coefficients.
A beta coefficient $\beta_i$ is a dimensionless quantity. Thus, we can directly compare beta coefficients.
The importance of a feature is evaluated by the absolute value of the beta coefficient. A larger value means greater importance.

Since the relationship between features and GLM outcome is simple and straightforward, we can interpret the model easily;
however, it is difficult to construct accurate prediction models that include interactions and non-linear terms using the GLM.

\subsection{Decision Tree and Ensemble Trees}
A decision tree is a classification algorithm that partitions data into subsets. The partitioning process is a binary split and the outcome is described by simple if-then rules.
The decision tree model is easy for interpretation; however, it is difficult to construct an accurate model using a single decision tree.

An ensemble tree combines several decision trees to achieve a better predictive performance than a single decision tree.
The main principle behind the ensemble model is that a group of weak learners comes together to form one strong learner.
The Random Forest and GBDT\cite{friedman2001greedy} are examples of ensemble tree models.
The ensemble tree model treats following data well.
\begin{itemize}
\item high dimensional features
\item non-linear relationships between features and outcome
\item interaction among features.
\end{itemize}
Since the model achieves good prediction performance empirically using high-dimensional big data, the ensemble model is widely used in the machine learning field.
The use of multiple decision trees can achieve a good prediction performance; however, interpretation of if-then rules in such a model is more difficult.

In the ensemble tree model, \textit{partial dependence plot}~\cite{friedman2001greedy,nohara2015interpreting} is used to show the relationship between the features and the outcome.

\subsection{Shapley Additive Explanation}
In the ensemble tree model, \textit{gain}, which is the total reduction of loss contributed by all splits for a given feature, is widely used to evaluate feature importance~\cite{friedman2001elements}; however, Lundberg et al. pointed out that gain is inconsistent.
This means that the importance of a given feature may decrease even if a model can change such that it relies more on the feature~\cite{lundberg2018consistent}.
To address these problems, Lundberg et al. proposed using the \textit{Shapley value} for calculating feature importance~\cite{lundberg2017unified}.

The Shapley value is a fair profit allocation among many stakeholders depending on their contribution~\cite{roth1988shapley} and was derived from the name of the economist who introduced it. The Shapley value is defined as follows:
\begin{equation}
\Phi(x_i)=\sum_{S \subseteq \{1,2,\cdots,K\} \setminus \{i\}} \frac{|S|!(K-|S|-1)!}{K!}[f_{x}(S\cup\{i\})-f_{x}(S)],
\label{SHAPdef}
\end{equation}
where $K$ is the number of stakeholders.
The meaning of the bracket part of Eq.~(\ref{SHAPdef}) is that the contribution of entitiy-$i$ can be defined as a marginal contribution, i.e. the difference between the profit obtained by group-$S$ members only: $f_{x}(S)$ and that of both entity-$i$ and the group members: $f_{x}(S\cup\{i\})$.
The gain introduced by Friedman~\cite{friedman2001elements} is based on the same idea; however, the gain can be changed by the group members $S$\footnote{In the tree model, group members correspond to all nodes from the root to the evaluating node.} and this causes inconsistent.
Then, we repeat this computation for all possible combinations and the Shapley value is the average of the marginal contributions of all possible combinations.
The Shapley value is the only profit allocation method that satisfies the following four properties: efficiency, symmetry, linearity and null player.
By using the idea of the Shapley value, \textit{SHapley Additive exPlanation} (SHAP) represents the outcome of patient-$j$: $f(x^{(j)})$ as the sum of each features-$i$'s contribution $\phi_i(x_i^{(j)})$.
\begin{equation}
\phi_0=\frac{1}{N} \sum_{j=1}^{N} f(x^{(j)})
\end{equation}
\begin{equation}
\phi_i(x_i^{(j)})=\Phi(x_i^{(j)}) - \frac{1}{N} \sum_{k=1}^{N} \Phi({x_i}^{(k)})
\label{phi}
\end{equation}
\begin{equation}
f(x^{(j)})=\phi_0+\sum_{i=1}^{K} \phi_i(x_i^{(j)})
\end{equation}
, where $N$ is the number of patients.
We derived  $\forall i$, $E(\phi(X_i))=\frac{1}{N} \sum_{j=1}^{N} \phi_i(x_i^{(j)})=0$ from Eq.(\ref{phi}).

The SHAP value has been proven to be consistent~\cite{lundberg2017unified} and is adoptable for all machine learning algorithms, including GLM.
The computation time of naive SHAP calculations increases exponentially with the number of features $K$; however, Lundberg et al. proposed polynomial time algorithm for decision trees and ensembles trees model~\cite{lundberg2018consistent}.
This algorithm is integrated into major ensemble tree frameworks like XGBoost~\cite{chen2016xgboost} and LightGBM~\cite{ke2017lightgbm}.

The relationship between a feature $x_i$ and SHAP value in GLM $\phi_{GLM}$ is given as follows~\cite{lundberg2017unified}:
\begin{equation}
\phi_{GLM}(x_i^{(j)}) =a_i x_i^{(j)} - E(a_i X_i)=a_i x_i^{(j)}-a_i E(X_i)
\label{phiGLM}
\end{equation}
Feature $x_i$ has a proportional relation with its SHAP value and the proportionality factor is given by coefficient $a_i$. This result is consistent with existing interpretations of GLM.

\section{Methods}

\subsection{SHAP Dependence Plot}
A \textit{SHAP dependence plot}~\cite{lundberg2018consistent} shows the relationship between the feature and its effect on the outcome measured by SHAP.
In binary prediction, SHAP values correspond to log-odds in the logistic regression model. 
The SHAP dependence plot for GLM shows the linear relationship given by Eq.~(\ref{phiGLM}).

\subsection{Variable Importance of SHAP}

Lundberg's original method~\cite{lundberg2018consistent} uses the sum of absolute value (L1-norm) of each patients' SHAP value to measure the feature's contribution:
$\sum_{j=1}^{N} | \phi(x_i^{(j)}) |$.
However, we propose using the variance (L2-norm) of the SHAP value for measuring variable importance.
\begin{eqnarray}
IMP(X_i)&=&Var(\phi(X_i)) =\frac{1}{N} \sum_{j=1}^{N} \left[\phi(x_i^{(j)})-E(\phi(X_i))\right]^2 \nonumber \\
%&=&\frac{1}{N} \sum_{j=1}^{N} \left[\{\phi(x_i^{(j)})\}^2 -2\phi(x_i^{(j)})E(\phi(X_i))\ + \{E(\phi(X_i))\}^2\right] \nonumber \\
%&=&\frac{1}{N} \sum_{j=1}^{N} \left[\phi(x_i^{(j)})\right]^2-\frac{2}{N} E(\phi(X_i)) \sum_{j=1}^{N} \phi(x_i^{(j)})+\{E(\phi(X_i))\}^2 \nonumber \\
&=&\frac{1}{N} \sum_{j=1}^{N} \left[\phi(x_i^{(j)})\right]^2-\{E(\phi(X_i))\}^2 \nonumber \\
&=&\frac{1}{N} \sum_{j=1}^{N} \left[\phi(x_i^{(j)})\right]^2
\label{imp}
\end{eqnarray}

Using our definition, the variable importance of GLM in Eq.~(\ref{glm}), in which all features are standardized, is given as follows:
\begin{eqnarray}
IMP_{GLM}(X_i)
&=& \frac{1}{N} \sum_{j=1}^{N} \left[a_i x_i^{(j)}-a_i E(X_i)\right]^2 \nonumber \\
&=& \frac{a_i^2}{N} \sum_{j=1}^{N} \left[x_i^{(j)}-E(X_i)\right]^2 \nonumber \\
&=& |a_i|^2 \cdot Var(X_i)=|\beta_i|^2
\end{eqnarray}
The ranking result sorted by the absolute value of the beta coefficients $\beta_i$ is exactly the same as that of our definition.
Our definition is also useful for the feature packing technique described in the next subsection.

\subsection{SHAP Feature Packing}

Explanatory variables often include similar features. For example, weight, height, and BMI are body-related information and correlated with each other.

Since correlated variables may prevent interpretation, it is common to remove correlated variables from the data and reconstruct the predictor.
However, this method requires more calculation time for reconstruction and the prediction accuracy may become worse.
Thus, we want to pack these variables into one grouped variable without reconstruction of the predictor.
Thanks to a characteristic of the SHAP value, the effect of two features on the outcome $\phi(\{x_i^{(j)},x_k^{(j)}\})$ is calculated as follows:
\begin{equation}
\phi(\{x_i^{(j)},x_k^{(j)}\})=\phi(x_i^{(j)})+\phi(x_k^{(j)})
\end{equation}
Next, we derive the feature importance of the grouped variable $IMP(\{X_i, X_k\})$ by the definition of Eq.~(\ref{imp}).
The grouped feature importance is given in a simple form as follows:
\begin{eqnarray}
\lefteqn{IMP(\{X_i, X_k\})} \nonumber \\
&=& Var(\phi(X_i)+\phi(X_k)) \nonumber \\
&=& Var(\phi(X_i))+Var(\phi(X_k))+2 \cdot Cov(\phi(X_i),\phi(X_k)) \nonumber \\
&=& IMP(X_i)+IMP(X_k)+2 \cdot Cov(\phi(X_i),\phi(X_k))
\end{eqnarray}
, where $Cov(X, Y)$ is the covariance between $X$ and $Y$.
The high covariance of SHAP values means that these features operate in a similar way for the outcome or there are strong interactions between them; thus, it is reasonable to pack these variables into one grouped variable.
The grouped feature importance is larger than the sum of each feature importance and a higher covariance derives a higher grouped importance.
Moreover, feature packing does not require reconstruction of the model and has absolutely no impact on prediction accuracy.
Therefore, packing of features with similar meanings or large covariate is a very useful technique.

\subsection{SHAP Summary Plot}
A \textit{SHAP summary plot}~\cite{lundberg2018consistent} shows the feature importance and a summary of the SHAP dependence plots.
In the plot, features are sorted by their importance, as defined by Eq.~(\ref{imp}), and stacked vertically.
Each row plot is a summary of the SHAP dependence plot of each feature $X_i$.
Each dot represents a patient's SHAP value $\phi(x_i^{(j)})$ plotted horizontally.
Each dot is colored by the value of the feature, from low (blue) to high (red).  Black dots represent missing values.
If red points are plotted at the lower side and blue dots are plotted at the higher side, then the risk becomes higher as the value increases.
Since a SHAP summary plot shows the importance of feature values and an abstract of the SHAP dependence plot, it is useful for overviewing the SHAP analysis.

\subsection{Flow of SHAP Analysis}
First, we provide an overview of the constructed model by the SHAP summary plot. 
We find interesting features from the summary plot, and draw SHAP dependence plots for detailed analysis.
If necessary, we identify similar features by their meaning or covariate of SHAP value, and pack these features into a grouped feature.

\section{Experiment and Results}

In total, 1712 patients were admitted to Saiseikai Kumamoto Hospital for cerebral infarction from October 2011 to October 2016 and applied clinical pathway for mild cerebral infarction is 1712. 
Of these, 1534 patients ($\sim$90\%) showed almost no significant disability at the preclinical stage; that is, their pre-onset modified ranking score (mRS) was less than 3.
Using data from these patients (N = 1534), we generated a predictor of whether their outcome worsened, that is, if discharge mRS was more than 2.
The number of positive patients was 262 of 1534.
Since one aim of this analysis was to identify high-risk patients in advance, we extracted data available on admission day from five data sources:
\begin{itemize}
\item \textbf{Standardized discharge summary: }
Many acute hospitals in Japan, including Saiseikai Kumamoto Hospital, participate DPC (Diagnosis Procedure Combination) payment system and create standardized DPC data in order to submit to the authorities.
DPC data includes structured discharge summary, known as Format-1~\cite{Yasunaga2019}, and we extracted data available on admission day from Format-1 data.
The following is an example:
age, height, weight, Japan Comma Scale (JCS) on admission, activity of daily living (ADL) on admission, smoking index, tPA, etc.
\item \textbf{Discharge summary from the neurology department:}
The neurology department create department-specific discharge summary.
The specific summary includes text documents and we eliminate them and extract numerical or categorical data from the specific summary. The following is an example:
NIH Stroke Scale (NIHSS), chief complaint, acute-phase treatment, Glasgow Coma Scale (GCS), onset time, admission time, etc.
\item \textbf{Plan of nutritional management: }
We extract numerical or categorical data from nutritional plan. The following is an example:
body mass index (BMI), basic energy expenditure (BEE), etc.
\item \textbf{Nursing Care:}
We extract numerical or categorical data from nursing care data. The following is an example:
observations (level of paralysis, pupil diameter, aching pain, etc.), vital signs (blood pressure, body temperature, oxygen saturation, etc.), dietary intake, etc.
\item \textbf{Examination outcome:}
We extract numerical or categorical data from examination outcome. The following is an example:
blood test (D-dimer, albumin, potassium, blood urea nitrogen, C-reactive protein, etc.), blood gas analysis (hemoglobin, red blood cell, etc.), urinary test (pH, urinary specific gravity, etc.)
\end{itemize}

All target patients are applied clinical pathway; however, analysis data does not include pathway items because we only used data at admission.
In addition, we convert a nominal data into multiple numerical features using one-hot encoding.
We then constructed a predictor of prognosis using GBDT. 
The model was analyzed by the existing method and the SHAP method. We then compared the results of these two methods.

A total of 1714 features were extracted from the 5 data sources available on the day of admission.
The mean of the cross-validated area under the curve (AUC) of the predictor was 0.788 and its standard deviation was 0.006.
This indicates that we achieved good prediction accuracy.

\subsection{Analysis by the Existing Method}
Before starting SHAP analysis, we evaluated the predictor using the existing method.

\begin{figure}[tb]
  \centering
  \includegraphics[width=\linewidth]{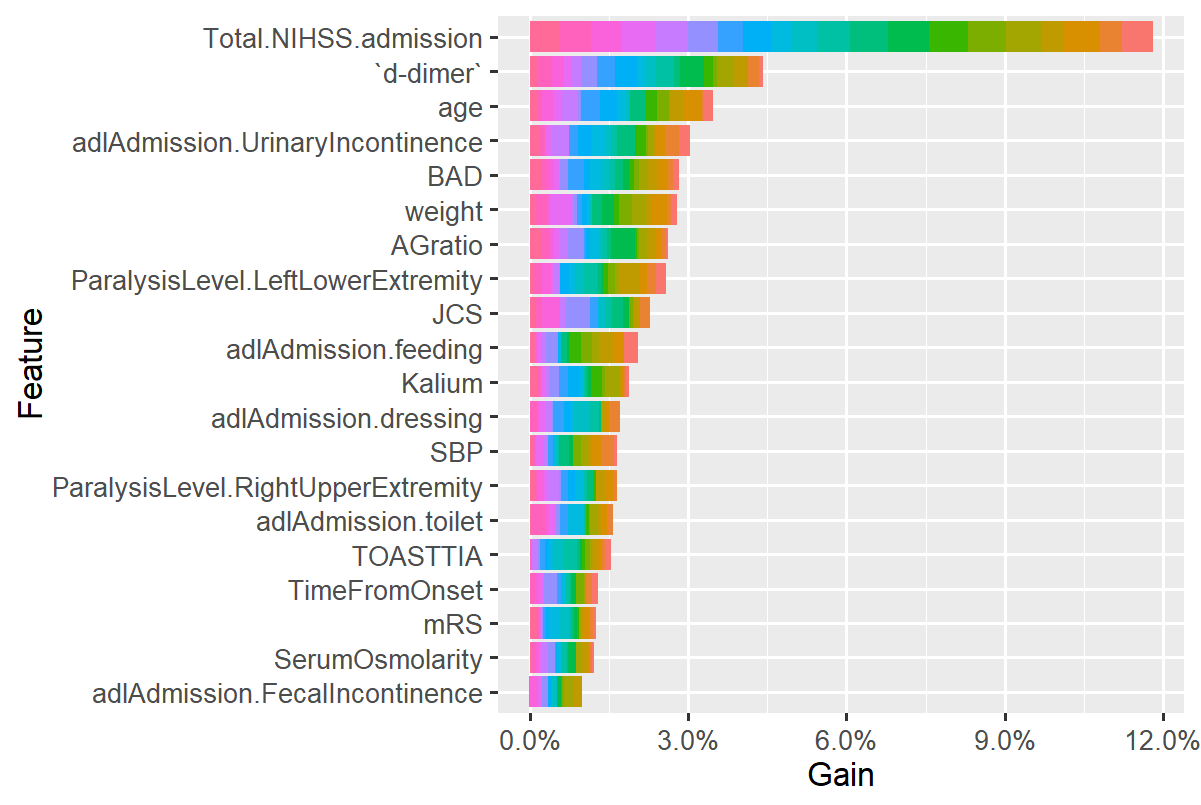}
  \caption{Feature Importance Plot: The top 20 important features are listed and sorted by the gain method. Total NIH Stroke Score on admission was extracted as the most important feature of the predictor by the gain method.}
  \label{gain}
\end{figure}

\begin{figure}[tb]
  \centering
  \includegraphics[width=\linewidth]{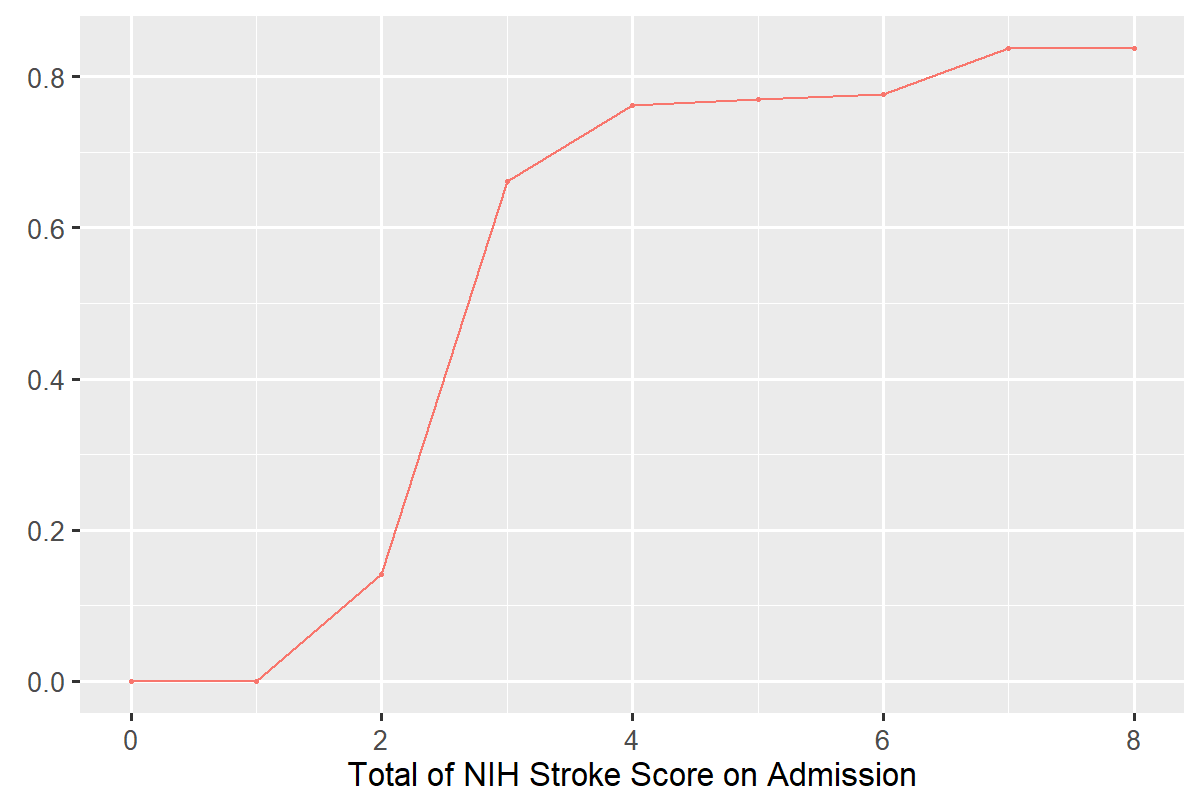}
  \caption{Partial dependence plot showing the relationship between total NIHSS and outcome: The plot shows that a higher total NIHSS derives a higher risk of poor prognosis.}
  \label{PDPnihss}
\end{figure}

Figure~\ref{gain} is a feature importance plot that shows the results of evaluating feature importance using the existing gain method.
The feature importance plot gives the relative importance, but it does not show the range and distribution of features or how the features relate to the outcome.
Next, we draw a partial dependence plot in Figure~\ref{PDPnihss} to show the relationship between total NIHSS and outcome.
The partial dependence plot shows that a higher total NIHSS derives a higher risk of bad prognosis and that the threshold for worse prognosis is three.

\subsection{Analysis by the SHAP Method}
We evaluated the predictor using the SHAP method and then compared the results between the existing method and the SHAP method.

\subsubsection{SHAP Summary Plot}

\begin{figure}[tb]
  \centering
  \includegraphics[width=\linewidth]{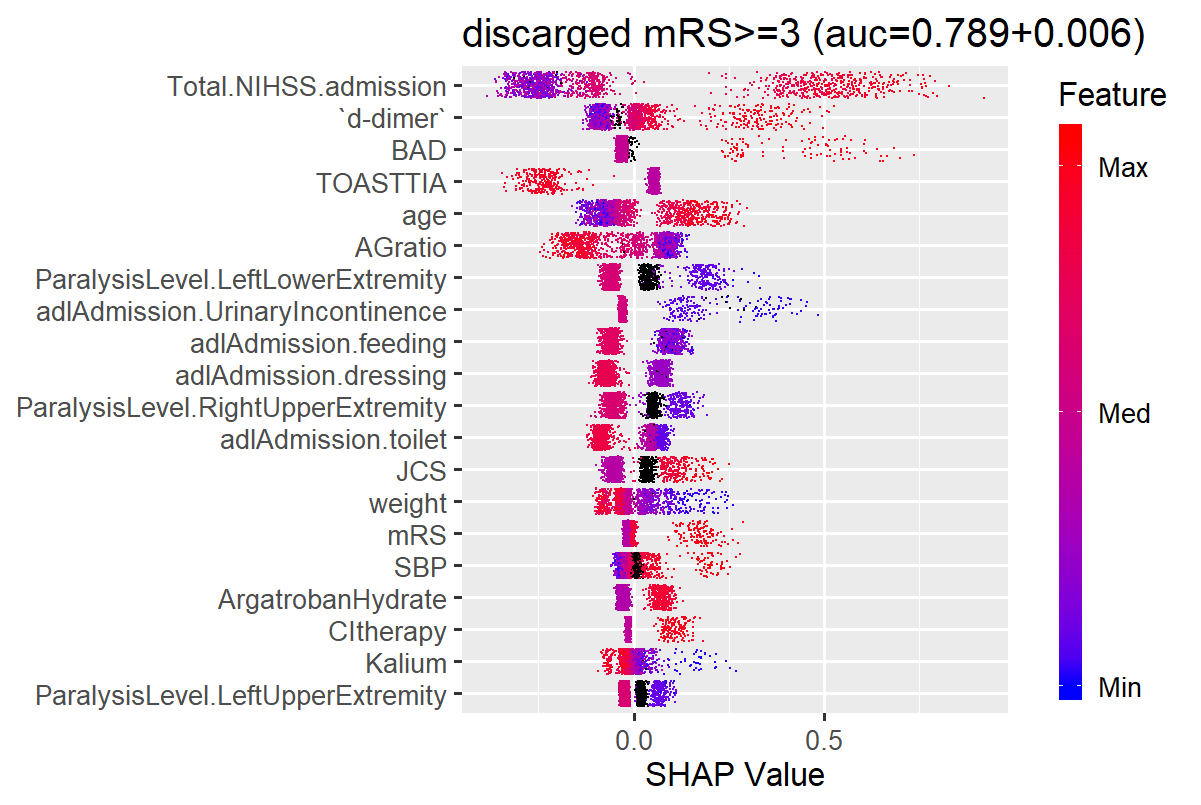}
  \caption{SHAP Summary Plot: The plot shows the top 20 important features evaluated by the SHAP method and the effects of each feature on the outcome. Total NIHSS on admission was extracted as the most important feature.}
  \label{SHAPsummary}
\end{figure}

Figure~\ref{SHAPsummary} presents the SHAP summary plot of the model.
Total NIHSS was also extracted as the most important feature of the predictor using the SHAP method.
We compared the top 20 important features between the existing method and the SHAP method. Although there were few changes in the ranking, 18 out of 20 features matched.
Lundberg et al. pointed out that the gain method is inconsistent and that it is impossible to reliably compare feature attribution values~\cite{lundberg2018consistent}.
In this analysis, there was no significant difference of ranking between the two methods.

On the contrary, it should be noted that the feature importance plot does not tell us the effect of total NIHSS; rather, the SHAP summary plot tell us that a larger NIHSS derives a higher risk because the red points are plotted at the lower side while the blue points are plotted at the higher side.
The summary plot also shows that the maximum difference of the SHAP value is approximately 0.8 on the log-odds scale and the maximum odds is estimated at approximately 2.2.
Thus, the SHAP summary plot is more useful than the feature importance plot for overviewing the results.

The top 20 important features extracted by the SHAP analysis include NIHSS, D-dimer, branch atheromatous disease (BAD), etc. While these features are consistent with clinicians' expectations, the albumin/globulin ratio (A/G ratio), which was extracted as the sixth important feature, is unexpected.

\subsubsection{SHAP Dependence Plot}
\begin{figure}[tb]
  \centering
  \includegraphics[width=\linewidth]{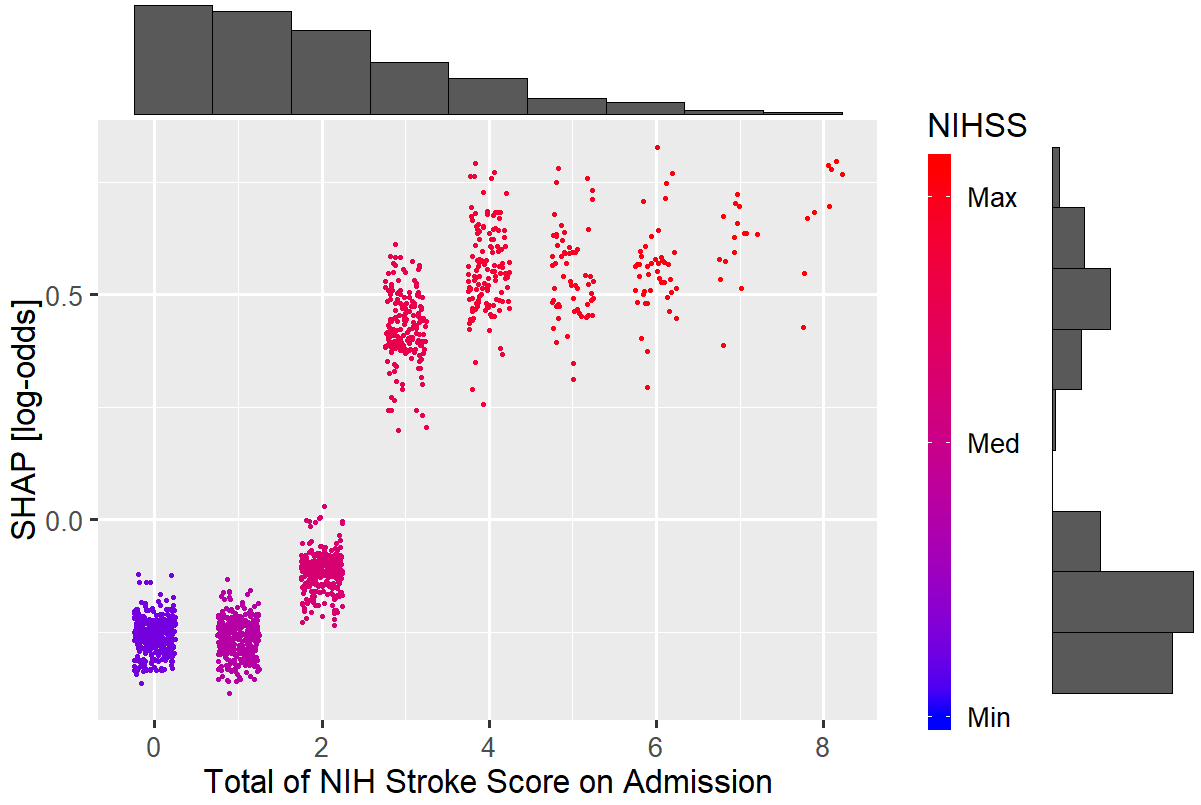}
  \caption{SHAP dependence plot showing the relationship between total NIHSS and outcome. The plot shows that a higher total NIHSS derives a higher risk of bad prognosis.}
  \label{SHAPnihss}
\end{figure}

For detailed analysis, we draw the SHAP dependence plot of total NIHSS in Figure~\ref{SHAPnihss}.
Like the partial dependence plot, the SHAP dependence plot shows that a higher total NIHSS derives  a higher risk of bad prognosis and that the threshold for worse prognosis is three.
In the SHAP dependence plot, the lines connected with the means of SHAP for each NIHSS are almost the same as those in the partial dependence plot.
This is because the partial dependence plot draws the mean effects of the estimated outcome if the feature is virtually changed.
The partial dependence plot only draws the mean effects and the variation in the estimation disappears; however, the SHAP dependence plot draws both the mean effects and variation.
This variation implies the existence of interaction. The  SHAP dependence plot is more informative than the partial dependence plot.

The graphs above and to the left of the SHAP dependence plot are the histograms of the X-axis and Y-axis, respectively.
Generally speaking, the prediction of the dense area is more accurate than that of the sparse area. The histograms help to understand the accuracy of the analysis.

\begin{figure}[tb]
  \centering
  \includegraphics[width=\linewidth]{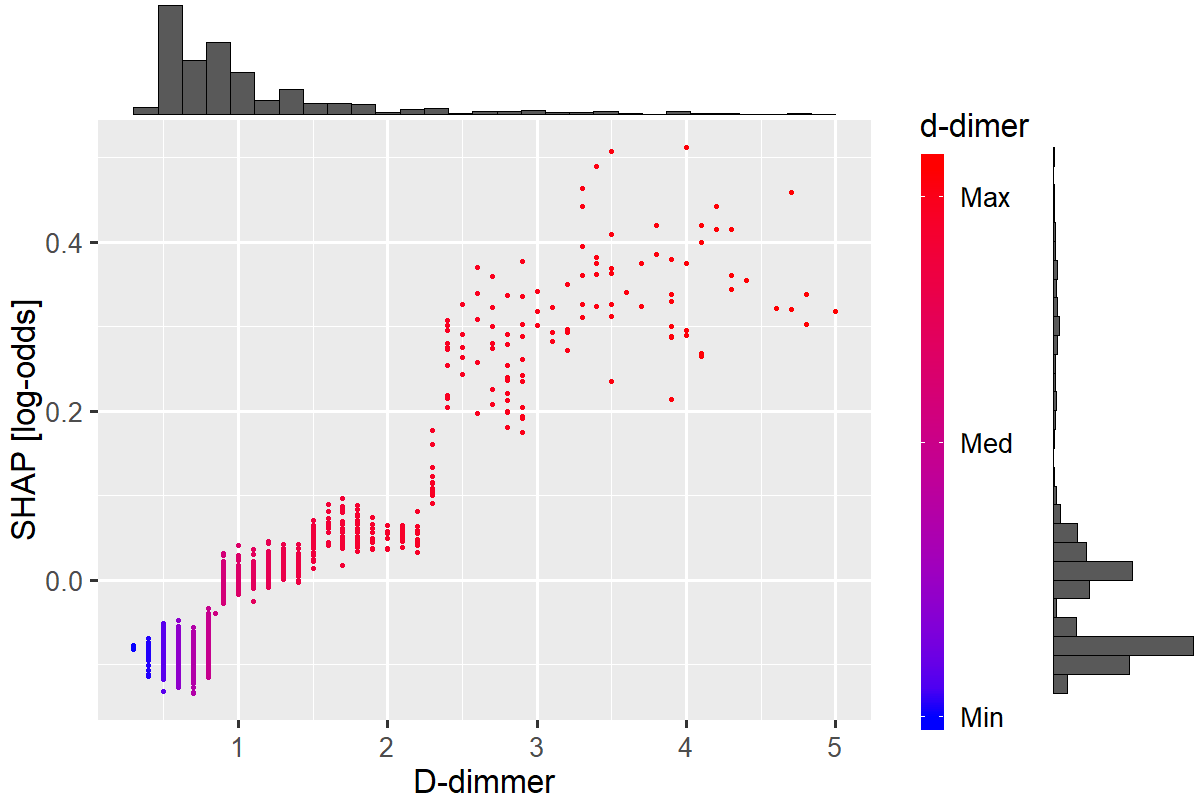}
  \caption{SHAP Dependence Plot of D-dimer.}
  \label{SHAPdimer}
\end{figure}

\begin{figure}[tb]
  \centering
  \includegraphics[width=\linewidth]{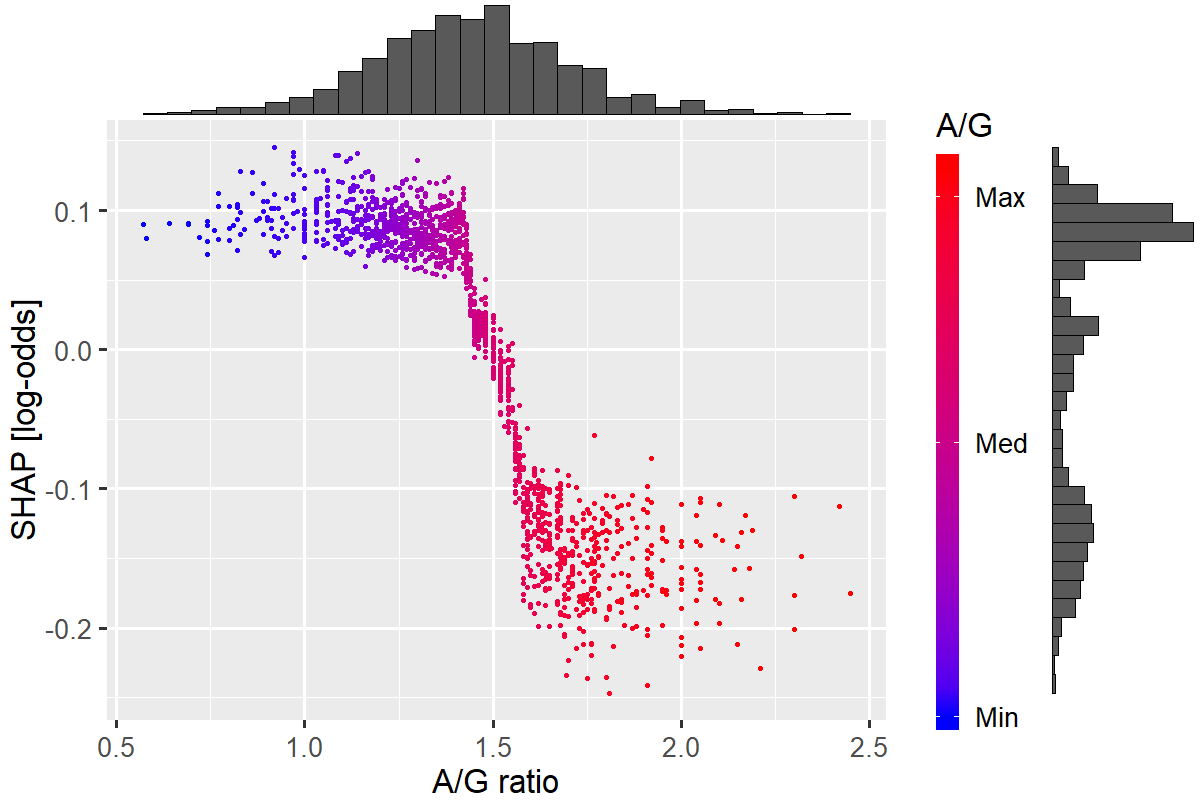}
  \caption{SHAP Dependence Plot of A/G Ratio.}
  \label{SHAPag}
\end{figure}

Figure~\ref{SHAPdimer} shows the SHAP dependence plot of D-dimer, the third important feature. The value of D-dimer has almost a proportional relationship with the risk of bad prognosis.
Figure~\ref{SHAPag} shows the SHAP dependence plot of A/G ratio, the eighth important feature and one that is unexpected by clinicians. The plot indicates that a value of 1.5 is the threshold of a good prognosis and the odds of patients with a high A/G ratio becomes 0.74[=exp(-0.3)].


\subsubsection{SHAP Feature Packing Technique}
\begin{figure}[tb]
  \centering
  \includegraphics[width=\linewidth]{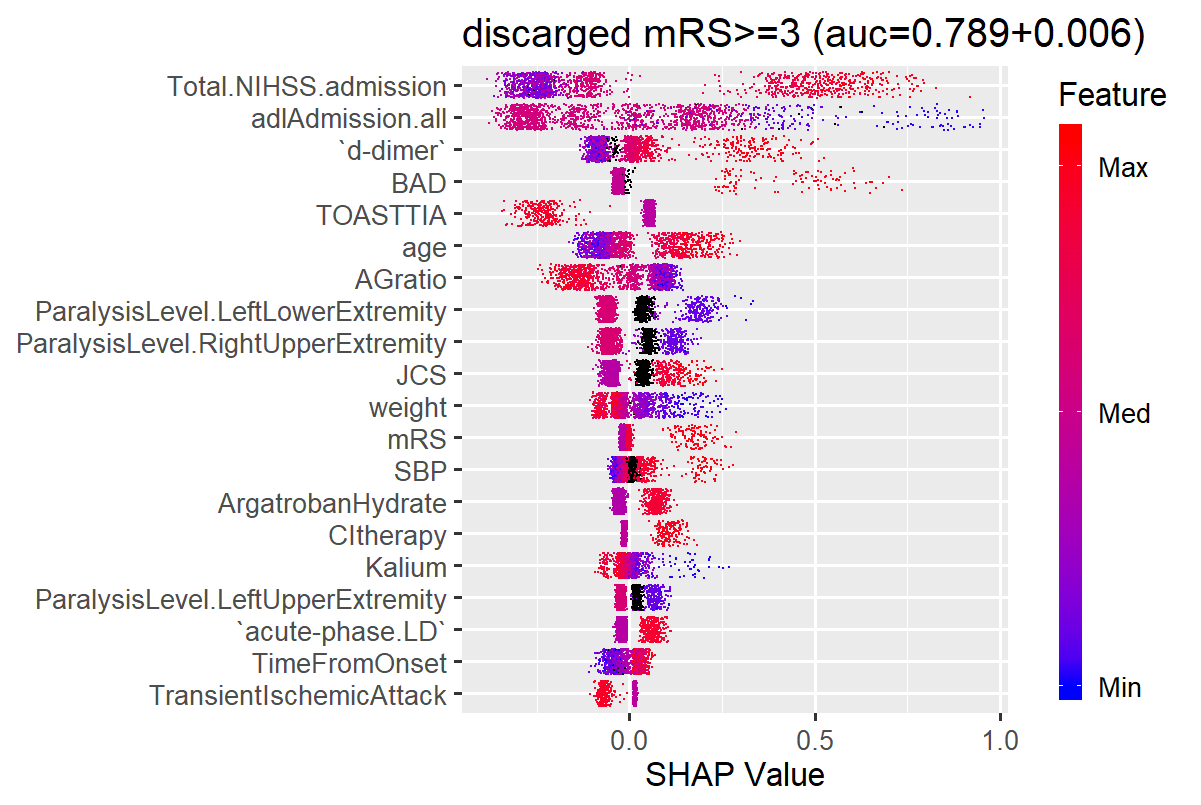}
  \caption{SHAP summary plot in which the 10 endpoints of ADL scores on admission are packed into one grouped variable named adlAdmission.all. adlAdmission.all became the second-most important features of the model.}
  \label{SHAPsummaryGroup}
\end{figure}

In the summary plot in Figure~\ref{SHAPsummary}, there are many features that start with `adlAdmission'.
These features are 10 endpoints of ADL scores on admission and are correlated with each other.
Packing these variables into one grouped variable helps to understand the model more easily.
We named the grouped variable `adlAdmission.all' and the result is shown in Figure~\ref{SHAPsummaryGroup}.
In Figure~\ref{SHAPsummary}, the importance of ADL scores are distributed to each endpoint. The integrated features become the second-most important features and are as important as NIHSS in Figure~\ref{SHAPsummaryGroup}.

\section{Discussion}

\subsection{Explanation of Missing Values}

In general machine learning methods, missing values cannot be handled directly and it is necessary to fill in the missing values with the median value, and so on.
By contrast, decision tree or ensemble tree models can handle missing values without data filling because the decision tree can learn whether missing values should branch to the right or left to increase accuracy. 
The SHAP value can be calculated for missing values. The SHAP plots show how missing values affect the outcome.

The effects of missing values depend on the generation mechanism of the missing values.
In the case of values missing at random (MAR), missing data are selected from both the high risk group and low risk group so that the impact on the outcome is considered to be intermediate between the two groups.
For example, since measurements of paralysis level, the seventh most important feature, and JCS level, the 13th most important feature, were performed for all stroke patients, in most cases the missing data occurred at random.
In such cases, the black dots representing the missing values for these features are plotted in the middle of the SHAP summary plot.

In other cases missing values may not occur randomly. For example, in our data, blood glucose levels were mainly measured for diabetic patients.
Most of the missing values are selected from patients with low glucose levels. The black dots may then be plotted at the right or left of the SHAP summary plot.

\subsection{Relationship between A/G ratio and outcome}
\begin{figure}[tb]
  \centering
  \includegraphics[width=\linewidth]{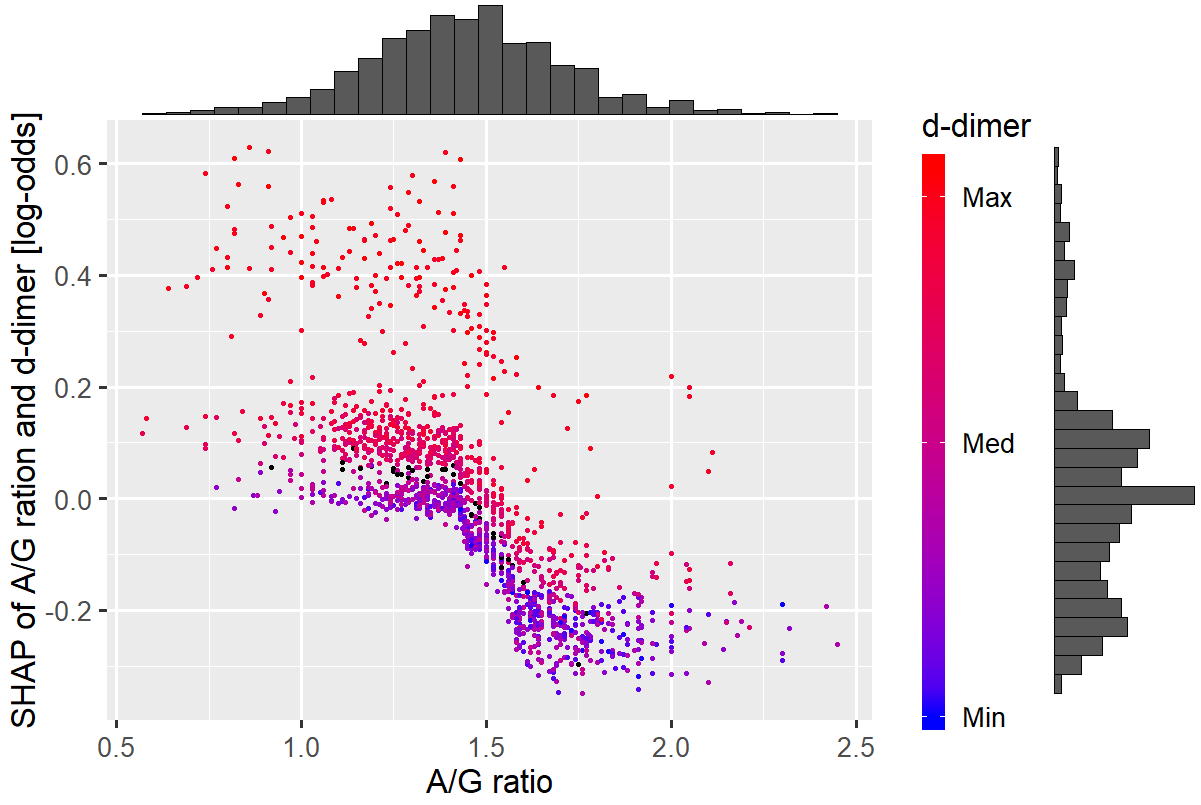}
  \caption{SHAP Dependence Plot of A/G ratio and D-dimer: the x-axis represents the A/G ratio and each dot is colored according to the patient's D-dimer.}
  \label{SHAPagGroup}
\end{figure}

The A/G ratio is the ratio of albumin to globulin in serum. A high A/G ratio group could mean a high albumin group or a low globulin group.
Belayev et al. reported that albumin has a brain protective effect for acute cerebral infarction and affects death and complications in the low albumin group~\cite{belayev1998diffusion}.
However, although albumin itself is included in the feature, its variable importance is low. Thus, it is unlikely that albumin is the only cause.
Although the mechanism is unknown, it is possible that globulin or some interaction term may affect the  A/G ratio.
We examined the largest covariance with the SHAP value of A/G ratio and found that the SHAP value of D-dimer is the largest one.
D-dimer is a degradation product of fibrin. A high D-dimer value indicates that a thrombus was formed recently.
D-dimer is known to be a risk marker for cerebral infarction~\cite{whiteley2008blood}. Figure~\ref{SHAPdimer} has already shown that patients with high D-dimer have a high risk of bad prognosis.
Therefore, we examined the effect of D-dimer and A/G ratio using the feature packing technique.
Figure~\ref{SHAPagGroup} shows that patients with high D-dimer (shown in red) were few among the groups whose A/G ratio is more than 1.5 and the most patients of high A/G ratio group belong to low D-dimer group (shown in blue).

\begin{figure}[tb]
  \centering
  \includegraphics[width=\linewidth]{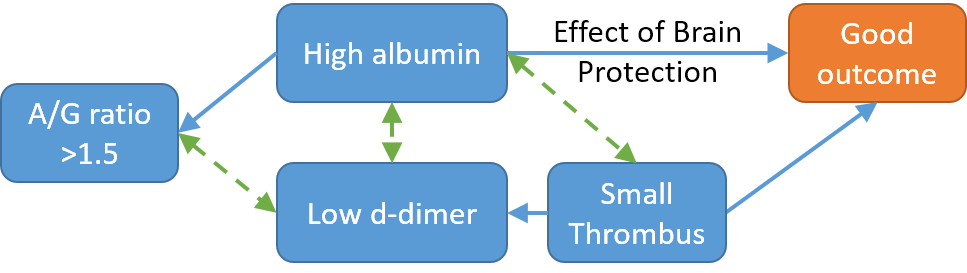}
  \caption{Estimated relationship between A/G ratio and patient outcome: Blue solid lines represent causal correlations and green dashed lines represent correlations.}
  \label{agRel}
\end{figure}

Figure~\ref{agRel} shows the estimated causal and correlation relationship. 
Since data analysis tells us only correlations, causal correlations are based by known facts.
Further analysis is necessary in the future.

\section{Conclusion}

In this paper, we adopted the SHAP method for interpreting GBDT model constructed using real data from our hospital.
For better interpretability, we proposed two novel techniques as follows: (1) a new metric of feature importance using SHAP and (2) a technique termed feature packing, which helps to understand the model more easily without reconstruction of the model.
We then compared the explanation results between the SHAP method and an existing method such as the gain feature importance and partial dependence plot.
The interpretation by SHAP was mostly consistent with that by the existing methods.
For overviewing the analysis results, the SHAP summary plot was more useful than the existing feature importance plots.
We also showed how the A/G ratio functions as an important prognosis factor for cerebral infarction using our hospital data and proposed techniques.
Our techniques are useful for interpreting machine learning models and can uncover the underlying relationships between features and outcome.

\section*{Acknowledgments}
This work was supported by JSPS KAKENHI Grant Number JP20K11938.

\section*{Conflict of Interest Statement}
The authors declare that there are no competing interests.

\bibliographystyle{elsarticle-num.bst}
\bibliography{SHAPreference}
\end{document}